\documentclass[11pt,a4paper]{article}
\usepackage[hyperref]{acl2019}
\usepackage{times}
\usepackage{latexsym}
\usepackage{graphicx}
\usepackage{multirow}
\usepackage{titlesec}
\usepackage{url}
\titlespacing{\subsection} {0pt}{9pt}{2pt}

\newcommand*\myat{{\fontfamily{ptm}\selectfont\small @}}

\aclfinalcopy 
\def\aclpaperid{1922} 

\title{Incremental Transformer with Deliberation Decoder \\ for Document Grounded Conversations}

\author{Zekang Li$^{\dagger\diamondsuit}$, Cheng Niu$^\ddagger$, Fandong Meng$^\ddagger$\thanks{$^*$ Fandong Meng is the corresponding author of the paper. This work was done when Zekang Li was interning at Pattern Recognition Center, WeChat AI, Tencent.}\ , Yang Feng$^\diamondsuit$, Qian Li$^\spadesuit$, Jie Zhou$^\ddagger$\\
  $^\dagger$Dian Group, School of Electronic Information and Communications \\ 
  Huazhong University of Science and Technology \\
  $^\ddagger$Pattern Recognition Center, WeChat AI, Tencent Inc, China \\
  $^\diamondsuit$Key Laboratory of Intelligent Information Processing \\
  Institute of Computing Technology, Chinese Academy of Sciences \\
  $^\spadesuit$School of Computer Science and Engineering, Northeastern University, China \\
  {\tt \small zekangli97@gmail.com, \{chengniu,fandongmeng,jiezhou\}@tencent.com} \\ {\tt \small fengyang\myat ict.ac.cn, qianli@stumail.neu.edu.cn}
  }

\date{}

\begin{document}
\maketitle

\begin{abstract}
Document Grounded Conversations is a task to generate dialogue responses when chatting about the content of a given document. Obviously, document knowledge plays a critical role in Document Grounded Conversations, while existing dialogue models do not exploit this kind of knowledge effectively enough. In this paper, we propose a novel Transformer-based architecture for multi-turn document grounded conversations. 
In particular, we devise an Incremental Transformer to encode multi-turn utterances along with knowledge in related documents. 
Motivated by the human cognitive process, we design a two-pass decoder (Deliberation Decoder) to improve context coherence and knowledge correctness.
Our empirical study on a real-world Document Grounded Dataset proves that responses generated by our model significantly outperform competitive baselines on both context coherence and knowledge relevance.  
\end{abstract}

\section{Introduction}

Past few years have witnessed the rapid development of dialogue systems.
Based on the sequence-to-sequence framework \cite{sutskever2014sequence}, most models are trained in an end-to-end manner with large corpora of human-to-human dialogues and have obtained impressive success \cite{shang2015neural, vinyals2015neural, li2015diversity, serban2016building}. 
While there is still a long way for reaching the ultimate goal of dialogue systems, which is to be able to talk like humans.
And one of the essential intelligence to achieve this goal is the ability to make use of knowledge.

There are several works on dialogue systems exploiting knowledge.
The Mem2Seq \cite{madotto2018mem2seq} incorporates structured knowledge into the end-to-end task-oriented dialogue. 
~\citeauthor{liu2018knowledge}~\shortcite{liu2018knowledge} introduces fact-matching and knowledge-diffusion to generate meaningful, diverse and natural responses using structured knowledge triplets. 
~\citeauthor{ghazvininejad2018knowledge}~\shortcite{ghazvininejad2018knowledge}, ~\citeauthor{parthasarathi2018extending}~\shortcite{parthasarathi2018extending}, ~\citeauthor{yavuzdeepcopy}~\shortcite{yavuzdeepcopy}, ~\citeauthor{dinan2018wizard}~\shortcite{dinan2018wizard} and ~\citeauthor{lo2019knowledge}~\shortcite{lo2019knowledge} apply unstructured text facts in open-domain dialogue systems. These works mainly focus on integrating factoid knowledge into dialogue systems, while factoid knowledge requires a lot of work to build up, and is only limited to expressing precise facts. Documents as a knowledge source provide a wide spectrum of knowledge, including but not limited to factoid, event updates, subjective opinion, etc. Recently, intensive research has been applied on using documents as knowledge sources for Question-Answering \cite{chen2017reading, huang2018flowqa, yu2018qanet, rajpurkar2018know, reddy2018coqa}. 

The Document Grounded Conversation is a task to generate natural dialogue responses when chatting about the content of a specific document. This task requires to integrate document knowledge with the multi-turn dialogue history. Different from previous knowledge grounded dialogue systems, Document Grounded Conversations utilize documents as the knowledge source, and hence are able to employ a wide spectrum of knowledge. And the Document Grounded Conversations is also different from document QA since the contextual consistent conversation response should be generated.
To address the Document Grounded Conversation task, it is important to:
1) Exploit document knowledge which are relevant to the conversation;
2) Develop a unified representation combining multi-turn utterances along with the relevant document knowledge.

In this paper, we propose a novel and effective Transformer-based \cite{vaswani2017attention} architecture for Document Grounded Conversations, named Incremental Transformer with Deliberation Decoder. The encoder employs a transformer architecture to incrementally encode multi-turn history utterances, and incorporate document knowledge into the the multi-turn context encoding process. The decoder is a two-pass decoder similar to the Deliberation Network in Neural Machine Translation \cite{xia2017deliberation}, which is designed to improve the context coherence and knowledge correctness of the responses. The first-pass decoder focuses on contextual coherence, while the second-pass decoder refines the result of the first-pass decoder by consulting the relevant document knowledge, and hence increases the knowledge relevance and correctness. This is motivated by human cognition process. In real-world human conversations, people usually first make a draft on how to respond the previous utterance, and then consummate the answer or even raise questions by consulting background knowledge.

We test the effectiveness of our proposed model on Document Grounded Conversations Dataset \cite{zhou2018dataset}. Experiment results show that our model is capable of generating responses of more context coherence and knowledge relevance. Sometimes document knowledge is even well used to guide the following conversations.  Both automatic and manual evaluations show that our model substantially outperforms the competitive baselines. 

Our contributions are as follows:
\begin{itemize}
    \item We build a novel Incremental Transformer to incrementally encode multi-turn utterances with document knowledge together.
    \item We are the first to apply a two-pass decoder to generate responses for document grounded conversations. Two decoders focus on context coherence and knowledge correctness respectively.
\end{itemize}

\section{Approach}
\subsection{Problem Statement}
Our goal is to incorporate the relevant document knowledge into multi-turn conversations. Formally, let $\mathbf{U}=\mathbf{u}^{(1)},...,\mathbf{u}^{(k)},...,\mathbf{u}^{(K)}$ be a whole conversation composed of $K$ utterances. We use $\mathbf{u}^{(k)} = u^{(k)}_{1},...,u^{(k)}_{i},...,u^{(k)}_{I}$ to denote the $k$-th utterance containing $I$ words, where $u^{(k)}_{i}$ denotes the $i$-th word in the $k$-th utterance. For each utterance $\mathbf{u}^{(k)}$, likewise, there is a specified relevant document $\mathbf{s}^{(k)} = s^{(k)}_{1},...,s^{(k)}_{j},...,s^{(k)}_{J}$, which represents the  document related to the $k$-th utterance containing $J$ words.
We define the document grounded conversations task as generating a response $\mathbf{u}^{(k+1)}$ given its related document $\mathbf{s}^{(k+1)}$ and previous $k$ utterances $\mathbf{U}_{\leq k}$ with related documents $\mathbf{S}_{\leq k}$, where $\mathbf{U}_{\leq k} = \mathbf{u}^{(1)},...,\mathbf{u}^{(k)}$ and $\mathbf{S}_{\leq k} = \mathbf{s}^{(1)},...,\mathbf{s}^{(k)}$. Note that $\mathbf{s}^{(k)}, \mathbf{s}^{(k+1)},...,\mathbf{s}^{(k+n)}$ may be the same.

Therefore, the probability to generate the response $\mathbf{u}^{(k+1)}$ is computed as:
\begin{equation}
    \begin{array}{ll}
         & P(\mathbf{u}^{(k+1)}|\mathbf{U}_{\leq k}, \mathbf{S}_{\leq k+1}; \theta)  \\
         & =\prod_{i=1}^{I} P(u^{k+1}_{i}|\mathbf{U}_{\leq k}, \mathbf{S}_{\leq k+1}, u^{(k+1)}_{<i};\theta)
    \end{array}
\end{equation}
where $u_{< i}^{(k+1)} = u_{1}^{(k+1)},...,u_{i-1}^{(k+1)}$.

\subsection{Model Description}
\begin{figure}[t]
\centering
\small
\includegraphics[width=0.48\textwidth]{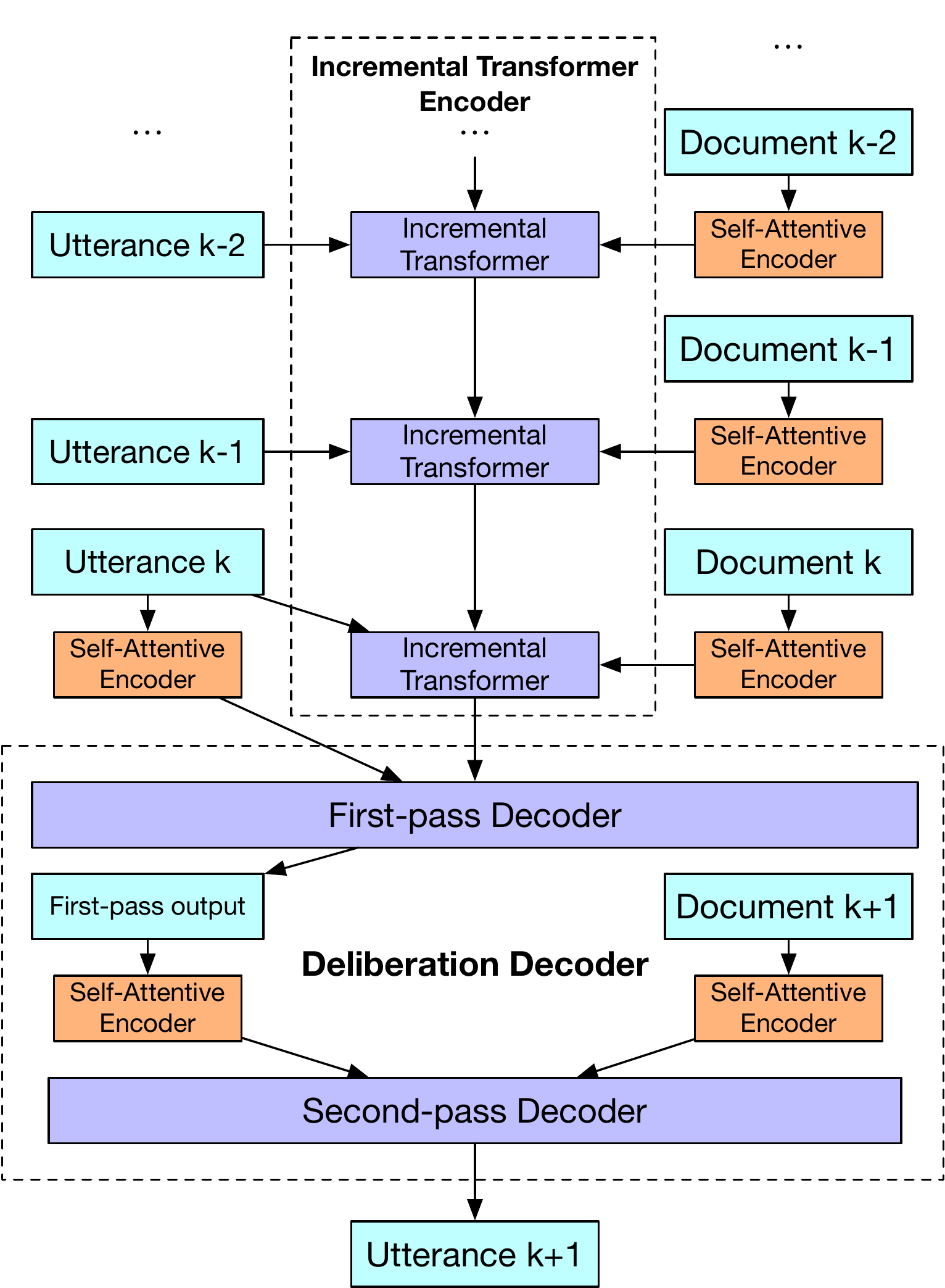}
\caption{The framework of Incremental Transformer with Deliberation Decoder for Document Grounded Conversations. } \vspace{-10pt}
\centering
\label{model architecture}
\end{figure}

\begin{figure*}[t]
\centering
\includegraphics[width=0.98\textwidth]{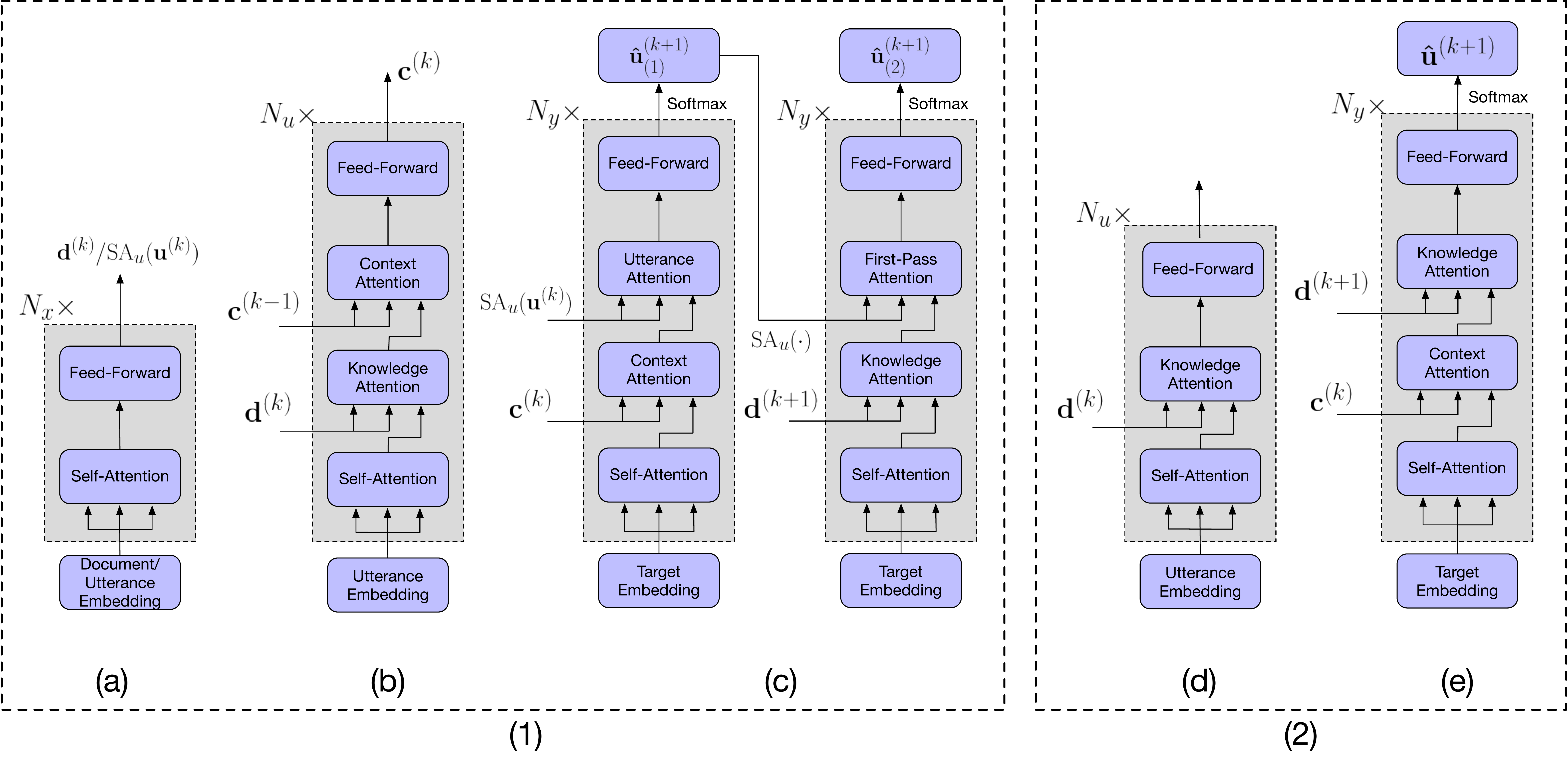}
\caption{(1) Detailed architecture of model components. (a) The Self-Attentive Encoder(SA). (b) Incremental Transformer (ITE).  (c) Deliberation Decoder (DD). (2) Simplified version of our proposed model used to verify the validity of our proposed Incremental Transformer Encoder and Deliberation Decoder. (d) Knowledge-Attention Transformer(KAT). (e) Context-Knowledge-Attention Decoder (CKAD).}  \vspace{-10pt}
\centering
\label{model modules}
\end{figure*}

Figure \ref{model architecture} shows the framework of the proposed Incremental Transformer with Deliberation Decoder. Please refer to Figure \ref{model modules} (1) for more details. It consists of three components: 

1) Self-Attentive Encoder (SA) (in orange) is a transformer encoder as described in \cite{vaswani2017attention}, which encodes the document knowledge and the current utterance independently.

2) Incremental Transformer Encoder (ITE) (on the top) is a unified transformer encoder which encodes multi-turn utterances with knowledge representation using an incremental encoding scheme. This module takes previous utterances $\mathbf{u}^{(i)}$ and the document $\mathbf{s}^{(i)}$'s SA representation as input, and use attention mechanism to incrementally build up the representation of relevant context and document knowledge.

3) Deliberation Decoder (DD) (on the bottom) is a two-pass unified transformer decoder for better generating the next response. The first-pass decoder takes current utterance $\mathbf{u}^{(k)}$'s SA representation and ITE output as input, and mainly relies on conversation context for response generation. The second-pass decoder takes the SA representation of the first pass result and the relevant document $\mathbf{s}^{(k+1)}$'s SA representation as input, and uses document knowledge to further refine the response.

\subsubsection*{Self-Attentive Encoder}

As document knowledge often includes several sentences, it's important to capture long-range dependencies and identify relevant information. 
We use multi-head self-attention \cite{vaswani2017attention} to compute the representation of document knowledge.

As shown in Figure \ref{model modules} (a), we use a self-attentive encoder to compute the representation of the related document knowledge $\mathbf{s}^{(k)}$. The input ($\mathbf{In}_{s}^{(k)}$) of the encoder is a sequence of document words embedding with positional encoding added.\cite{vaswani2017attention}: 
\begin{equation}
    \setlength{\abovedisplayskip}{4pt} 
    \setlength{\belowdisplayskip}{4pt}
    \mathbf{In}_{s}^{(k)} = [\mathbf{s}^{(k)}_{1},...,\mathbf{s}^{(k)}_{J}]
\end{equation}
\begin{equation}
    \setlength{\abovedisplayskip}{4pt} 
    \setlength{\belowdisplayskip}{4pt}
\mathbf{s}^{(k)}_{j} = \mathbf{e}_{s_{j}} + {\rm PE}(j)
\end{equation}
where $\mathbf{e}_{s_{j}}$ is the word embedding of $s_{j}^{(k)}$ and ${\rm PE}(\cdot)$ denotes positional encoding function.

The Self-Attentive encoder contains a stack of $N_{x}$ identical layers. Each layer has two sub-layers. The first sub-layer is a multi-head self-attention ($\rm MultiHead$) \cite{vaswani2017attention}. ${\rm MultiHead}(\mathbf{Q}, \mathbf{K}, \mathbf{V})$ is a multi-head attention function that takes a query matrix $\mathbf{Q}$, a key matrix $\mathbf{K}$, and a value matrix $\mathbf{V}$ as input. In current case, $\mathbf{Q}$ = $\mathbf{K}$ = $\mathbf{V}$. That's why it's called self-attention. And the second sub-layer is a simple, position-wise fully connected feed-forward network ($\rm FFN$). This $\rm FFN$ consists of two linear transformations with a ReLU activation in between. \cite{vaswani2017attention}.
\begin{equation}
    \setlength{\abovedisplayskip}{4pt} 
    \setlength{\belowdisplayskip}{4pt}
    \mathbf{A}^{(1)} = {\rm MultiHead}(\mathbf{In}_{s}^{(k)},\mathbf{In}_{s}^{(k)},\mathbf{In}_{s}^{(k)})
\end{equation}
\begin{equation}
    \setlength{\abovedisplayskip}{4pt} 
    \setlength{\belowdisplayskip}{4pt}
    \mathbf{D}^{(1)} = {\rm FFN}(\mathbf{A}^{(1)})
\end{equation}
\begin{equation}
    \setlength{\abovedisplayskip}{4pt} 
    \setlength{\belowdisplayskip}{4pt}
    {\rm FFN}(x) = \max (0, xW_{1}+b_{1})W_{2} + b_{2}
\end{equation}
where $\mathbf{A}^{(1)}$ is the hidden state computed by multi-head attention at the first layer, $\mathbf{D}^{(1)}$ denotes the representation of $\mathbf{s}^{(k)}$ after the first layer. Note that residual connection and layer normalization are used in each sub-layer, which are omitted in the presentation for simplicity. Please refer to \cite{vaswani2017attention} for more details.

For each layer, repeat this process:
\begin{equation}
\setlength{\abovedisplayskip}{4pt} 
    \setlength{\belowdisplayskip}{4pt}
    \mathbf{A}^{(n)} = {\rm MultiHead}(\mathbf{D}^{(n-1)},\mathbf{D}^{(n-1)},\mathbf{D}^{(n-1)})
\end{equation}
\begin{equation}
\setlength{\abovedisplayskip}{4pt} 
    \setlength{\belowdisplayskip}{4pt}
    \mathbf{D}^{(n)} = {\rm FFN}(\mathbf{A}^{(n)})
\end{equation}
where $n=1,...,N_{s}$ and $\mathbf{D}^{(0)} = \mathbf{In}_{s}^{(k)}$.

We use ${\rm SA}_{s}(\cdot)$ to denote this whole process:
\begin{equation}
    \setlength{\abovedisplayskip}{4pt} 
    \setlength{\belowdisplayskip}{4pt}
    \mathbf{d}^{(k)} = \mathbf{D}^{(N_{x})} = {\rm SA}_{s}(\mathbf{s}^{(k)})
\end{equation}
where $\mathbf{d}^{(k)}$ is the final representation for the document knowledge $\mathbf{s}^{(k)}$. 

Similarly, for each utterance $\mathbf{u}^{(k)}$, we use ${\mathbf{In}}_{u}^{(k)} = [\mathbf{u}^{(k)}_{1},...,\mathbf{u}^{(k)}_{I}]$ to represent the sequence of the position-aware word embedding. Then the same Self-Attentive Encoder is used to compute the representation of current utterance $\mathbf{u}^{(k)}$, and we use ${\rm SA}_{u}(\mathbf{u}^{(k)})$ to denote this encoding result. The Self-Attentive Encoder is also used to encode the document ${\mathbf{s}}^{(k+1)}$ and the first pass decoding results in the second pass of the decoder.  Note that ${\rm SA}_{s}$ and ${\rm SA}_{u}$ have the same architecture but different parameters. More details about this will be mentioned in the following sections.

\subsubsection*{Incremental Transformer Encoder}
To encode multi-turn document grounded utterances effectively, we design an Incremental Transformer Encoder.
Incremental Transformer uses multi-head attention to incorporate document knowledge and context into the current utterance's encoding process.
This process can be stated recursively as follows:
\begin{equation}
\setlength{\abovedisplayskip}{4pt} 
    \setlength{\belowdisplayskip}{4pt}
    \mathbf{c}^{(k)} = {\rm ITE}(\mathbf{c}^{(k-1)}, \mathbf{d}^{(k)}, \mathbf{In}_{u}^{(k)})
\end{equation}
where ${\rm ITE}(\cdot)$ denotes the encoding function, $\mathbf{c}^{(k)}$ denotes the context state after encoding utterance $\mathbf{u}^{(k)}$, $\mathbf{c}^{(k-1)}$ is the context state after encoding last utterance $\mathbf{u}^{(k-1)}$, $\mathbf{d}^{(k)}$ is the representation of document $\mathbf{s}^{(k)}$ and $\mathbf{In}_{u}^{(k)}$ is the embedding of current utterance $\mathbf{u}^{(k)}$.

As shown in Figure \ref{model modules} (b), we use a stack of $N_{u}$ identical layers to encode $\mathbf{u}^{(k)}$. Each layer consists of four sub-layers. 
The first sub-layer is a multi-head self-attention:
\begin{equation}
\setlength{\abovedisplayskip}{4pt} 
    \setlength{\belowdisplayskip}{4pt}
    \mathbf{B}^{(n)} = {\rm MultiHead}(\mathbf{C}^{(n-1)}, \mathbf{C}^{(n-1)}, \mathbf{C}^{(n-1)})
\end{equation}
where $n = 1,...,N_{u}$, $\mathbf{C}^{(n-1)}$ is the output of the last layer and $\mathbf{C}^{(0)} = \mathbf{In}_{u}^{(k)}$. The second sub-layer is a multi-head knowledge attention:
\begin{equation}
    \setlength{\abovedisplayskip}{4pt} 
    \setlength{\belowdisplayskip}{4pt}
    \mathbf{E}^{(n)} = {\rm MultiHead}(\mathbf{B}^{(n)}, \mathbf{d}^{(k)}, \mathbf{d}^{(k)})
\end{equation}
The third sub-layer is a multi-head context attention:
\begin{equation}
    \setlength{\abovedisplayskip}{4pt} 
    \setlength{\belowdisplayskip}{4pt}
    \mathbf{F}^{(n)} = {\rm MultiHead}(\mathbf{E}^{(n)}, \mathbf{c}^{(k-1)}, \mathbf{c}^{(k-1)})
\end{equation}
where $\mathbf{c}^{(k-1)}$ is the representation of the previous utterances. That's why we called the encoder "Incremental Transformer". The fourth sub-layer is a position-wise fully connected feed-forward network:
\begin{equation}
    \setlength{\abovedisplayskip}{4pt} 
    \setlength{\belowdisplayskip}{4pt}
    \mathbf{C}^{(n)} = {\rm FFN}(\mathbf{F}^{(n)})
\end{equation}
We use $\mathbf{c}^{(k)}$ to denote the final representation at $N_{u}$-th layer:
\begin{equation}
    \setlength{\abovedisplayskip}{4pt} 
    \setlength{\belowdisplayskip}{4pt}
    \mathbf{c}^{(k)} = \mathbf{C}^{(N_{u})}
\end{equation}

\subsubsection*{Deliberation Decoder}
Motivated by the real-world human cognitive process, we design a Deliberation Decoder containing two decoding passes to improve the knowledge relevance and context coherence. The first-pass decoder takes the representation of current utterance ${\rm SA}_{u}(\mathbf{\mathbf{u}^{(k)}})$ and context $\mathbf{c}^{(k)}$ as input and focuses on how to generate responses contextual coherently. The second-pass decoder takes the representation of the first-pass decoding results and related document $\mathbf{s}^{(k+1)}$ as input and focuses on increasing knowledge usage and guiding the following conversations within the scope of the given document.

When generating the $i$-th response word ${u}^{(k+1)}_{i}$, we have the generated words $\mathbf{u}^{(k+1)}_{<i}$ as input \cite{vaswani2017attention}. We use $\mathbf{In}_{r}^{(k+1)}$ to denote the matrix representation of $\mathbf{u}^{(k+1)}_{<i}$ as following:
\begin{equation}
\setlength{\abovedisplayskip}{4pt} 
    \setlength{\belowdisplayskip}{4pt}
    \mathbf{In}_{r}^{(k+1)} = [\mathbf{u}^{(k+1)}_{0}, \mathbf{u}^{(k+1)}_{1},...,\mathbf{u}^{(k+1)}_{i-1}]
\end{equation}
where $\mathbf{u}^{(k+1)}_{0}$ is the vector representation of sentence-start token. 

As shown in Figure \ref{model modules} (c), the Deliberation Decoder consists of a first-pass decoder and a second-pass decoder. These two decoders have the same architecture but different input for sub-layers. Both decoders are composed of a stack of $N_{y}$ identical layers. Each layer has four sub-layers.
For the first-pass decoder, the first sub-layer is a multi-head self-attention:
\begin{equation}
\setlength{\abovedisplayskip}{4pt} 
    \setlength{\belowdisplayskip}{4pt}
    \mathbf{G}^{(n)}_{1} = {\rm MultiHead}(\mathbf{R}^{(n-1)}_{1}, \mathbf{R}^{(n-1)}_{1}, \mathbf{R}^{(n-1)}_{1})
\end{equation}
where $n = 1,...,N_{y}$, $\mathbf{R}^{(n-1)}_{1}$ is the output of the previous layer, and $\mathbf{R}^{(0)}_{1} = \mathbf{In}_{r}^{(k+1)}$. The second sub-layer is a multi-head context attention:
\begin{equation} \label{eq18}
\setlength{\abovedisplayskip}{4pt} 
    \setlength{\belowdisplayskip}{4pt}
    \mathbf{H}^{(n)}_{1} = {\rm MultiHead}(\mathbf{G}^{(n)}_{1}, \mathbf{c}^{(k)}, \mathbf{c}^{(k)})
\end{equation}
where $\mathbf{c}^{(k)}$ is the representation of context $\mathbf{u}_{\leq k}$. The third sub-layer is a multi-head utterance attention:
\begin{equation} \label{eq19}
\setlength{\abovedisplayskip}{4pt} 
    \setlength{\belowdisplayskip}{4pt}
\begin{array}{cc}
     \mathbf{M}^{(n)}_{1} = {\rm MultiHead}(\mathbf{H}^{(n)}_{1}, & {\rm SA}_{u}(\mathbf{u}^{(k)}), \\
     & {\rm SA}_{u}(\mathbf{u}^{(k)}))
\end{array}
\end{equation}
where $SA_{u}(\cdot)$ is a Self-Attentive Encoder which encodes latest utterance $\mathbf{u}^{(k)}$. Eq. (\ref{eq18}) mainly encodes the context and document knowledge relevant to the latest utterance, while Eq. (\ref{eq19}) encodes the latest utterance directly. We hope optimal performance can be achieved by combining both. 

The fourth sub-layer is a position-wise fully connected feed-forward network:
\begin{equation}
    \setlength{\abovedisplayskip}{4pt} 
    \setlength{\belowdisplayskip}{4pt}
    \mathbf{R}^{(n)}_{1} = {\rm FFN}(\mathbf{M}^{(n)}_{1})
\end{equation}
After $N_{y}$ layers, we use softmax to get the words probabilities decoded by first-pass decoder:
\begin{equation}
    \setlength{\abovedisplayskip}{4pt} 
    \setlength{\belowdisplayskip}{4pt}
    P(\mathbf{\hat{u}}^{(k+1)}_{(1)})= {\rm softmax}(\mathbf{R}^{(N_{y})}_{1})
\end{equation}
where $\mathbf{\hat{u}}^{(k+1)}_{(1)}$ is the response decoded by the first-pass decoder.
For second-pass decoder:
\begin{equation}
\setlength{\abovedisplayskip}{4pt} 
    \setlength{\belowdisplayskip}{4pt}
    \mathbf{G}^{(n)}_{2} = {\rm MultiHead}(\mathbf{R}^{(n-1)}_{2}, \mathbf{R}^{(n-1)}_{2}, \mathbf{R}^{(n-1)}_{2})
\end{equation}
\begin{equation}
\setlength{\abovedisplayskip}{4pt} 
    \setlength{\belowdisplayskip}{4pt}
    \mathbf{H}^{(n)}_{2} = {\rm MultiHead}(\mathbf{G}^{(n)}_{2}, \mathbf{d}^{(k+1)}, \mathbf{d}^{(k+1)})
\end{equation}
\begin{equation}
\setlength{\abovedisplayskip}{4pt} 
    \setlength{\belowdisplayskip}{4pt}
\begin{array}{cc}
     \mathbf{M}^{(n)}_{2} = {\rm MultiHead}(\mathbf{H}^{(n)}_{2}, & {\rm SA}_{u}(\mathbf{\hat{u}}^{(k+1)}_{(1)}), \\
     & {\rm SA}_{u}(\mathbf{\hat{u}}^{(k+1)}_{(1)}))
\end{array}
\end{equation}
\begin{equation}
\setlength{\abovedisplayskip}{4pt} 
    \setlength{\belowdisplayskip}{4pt}
    \mathbf{R}^{(n)}_{2} = {\rm FFN}(\mathbf{M}^{(n)}_{2})
\end{equation}
\begin{equation}
\setlength{\abovedisplayskip}{4pt} 
    \setlength{\belowdisplayskip}{4pt}
    P(\mathbf{\hat{u}}^{(k+1)}_{(2)}) = {\rm softmax}(\mathbf{R}^{(N_{y})}_{2})
\end{equation}
where $\mathbf{R}^{(n-1)}_{2}$ is the counterpart to $\mathbf{R}^{(n-1)}_{1}$ in pass two decoder, referring to the output of the previous layer. $\mathbf{d}^{(k+1)}$ is the representation of  document $\mathbf{s}^{(k+1)}$ using Self-Attentive Encoder, $\mathbf{\hat{u}}^{(k+1)}_{(2)}$ is the output words after the second-pass decoder.

\subsubsection*{Training}

In contrast to the original Deliberation Network \cite{xia2017deliberation}, where they propose a complex joint learning framework using Monte Carlo Method, 
we minimize the following loss as ~\citeauthor{xiong2018modeling}~\shortcite{xiong2018modeling} do:
\begin{equation}
\setlength{\abovedisplayskip}{4pt} 
    \setlength{\belowdisplayskip}{4pt}
    L_{mle} = L_{mle1} + L_{mle2}
\end{equation}
\begin{equation}
\setlength{\abovedisplayskip}{4pt} 
    \setlength{\belowdisplayskip}{4pt}
    L_{mle1} = - \sum_{k=1}^{K}\sum_{i=1}^{I}\log P(\mathbf{\hat{u}}^{(k+1)}_{(1)i})
\end{equation}
\begin{equation}
\setlength{\abovedisplayskip}{4pt} 
    \setlength{\belowdisplayskip}{4pt}
    L_{mle2} = - \sum_{k=1}^{K}\sum_{i=1}^{I}\log P(\mathbf{\hat{u}}^{(k+1)}_{(2)i})
\end{equation}

\section{Experiments}
\subsection{Dataset}
We evaluate our model using the Document Grounded Conversations Dataset \cite{zhou2018dataset}. There are 72922 utterances for training, 3626 utterances for validation and 11577 utterances for testing. The utterances can be either casual chats or document grounded. Note that we consider consequent utterances of the same person as one utterance. For example, we consider \textit{A: Hello! B: Hi! B: How's it going?} as \textit{A: Hello! B: Hi! How's it going?}.
And there is a related document given for every several consequent utterances, which may contain movie name, casts, introduction, ratings, and some scenes. The average length of documents is about 200.
Please refer to \cite{zhou2018dataset} for more details.

\begin{table*}[!t]
\centering
\scalebox{0.95} {
\begin{tabular}{lccccc}
  & & &  & Knowledge  & Context \\
  Model & PPL & BLEU(\%) & Fluency & Relevance & Coherence\\
  \hline
  Seq2Seq without knowledge & 80.93 & 0.38 & 1.62 & 0.18 &  0.54\\
  HRED without knowledge & 80.84 & 0.43 & 1.25 & 0.18 &  0.30\\
  Transformer without knowledge & 87.32 & 0.36 & 1.60 & 0.29 & 0.67\\
  Seq2Seq (+knowledge) & 78.47 & 0.39 & 1.50 & 0.22 & 0.61\\
  HRED (+knowledge) & 79.12 & 0.77 & 1.56 & 0.35 & 0.47\\
  Wizard Transformer & 70.30 & 0.66 & 1.62 & 0.47 & 0.56\\
  \hline
  \textbf{ITE+DD} (ours) & \textbf{15.11} & \textbf{0.95} & 1.67 & \textbf{0.56}  & \textbf{0.90}\\
  \textbf{ITE+CKAD} (ours) & 64.97 & 0.86 & \textbf{1.68} & 0.50  & 0.82\\
  \textbf{KAT} (ours) & 65.36 & 0.58 & 1.58 & 0.33 & 0.78\\
  \hline
\end{tabular}
}
\caption{Automatic evaluation and manual evaluation results for baselines and our proposed models.} \vspace{-10pt}
\label{Automatic Evaluation and Munual Evaluation}
\end{table*}

\subsection{Baselines}
We compare our proposed model with the following state-of-the-art baselines:\\
\textbf{Models not using document knowledge}:

    \textbf{Seq2Seq}: A simple encoder-decoder model \cite{shang2015neural, vinyals2015neural} with global attention \cite{luong2015effective}. We concatenate utterances context to a long sentence as input.
    
    \textbf{HRED}: A hierarchical encoder-decoder model \cite{serban2016building}, which is composed of a word-level LSTM for each sentence and a sentence-level LSTM connecting utterances.
    
    \textbf{Transformer}: The state-of-the-art NMT model based on multi-head attention \cite{vaswani2017attention}. We concatenate utterances context to a long sentence as its input. \\
\textbf{Models using document knowledge}:

    \textbf{Seq2Seq (+knowledge)} and \textbf{HRED (+knowledge)} are based on Seq2Seq and HRED respectively. They both concatenate document knowledge representation and last decoding output embedding as input when decoding. Please refer to \cite{zhou2018dataset} for more details.
    
    \textbf{Wizard Transformer}: A Transformer-based model for multi-turn open-domain dialogue with unstructured text facts \cite{dinan2018wizard}. It concatenates context utterances and text facts to a long sequence as input. We replace the text facts with document knowledge.
    
Here, we also conduct an ablation study to illustrate the validity of our proposed Incremental Transformer Encoder and Deliberation Decoder.

    \textbf{ITE+CKAD}: It uses Incremental Transformer Encoder (ITE) as encoder and Context-Knowledge-Attention Decoder (CKAD) as shown in Figure \ref{model modules} (e). This setup is to test the validity of the deliberation decoder.
    
    \textbf{Knowledge-Attention Transformer (KAT)}: As shown in Figure \ref{model modules} (d), the encoder of this model is a simplified version of Incremental Transformer Encoder (ITE), which doesn't have context-attention sub-layer. We concatenate utterances context to a long sentence as its input. The decoder of the model is a simplified Context-Knowledge-Attention Decoder (CKAD). It doesn't have context-attention sub-layer either. This setup is to test how effective the context has been exploited in the full model.

\begin{table}[t]
    \centering
    \scalebox{0.95} {
    \begin{tabular}{|c|c|c|}
    \hline
         & Knowledge  &  Context \\
        Model & Relevance(\%) & Coherence(\%)\\ \hline
        Wizard & 64/25/11 & 58/28/14\\
        \hline
        \textbf{ITE+CKAD} & 67/16/17 & 40/37/23 \\ \hline
        \textbf{ITE+DD} & 64/16/20 & 38/34/28 \\ \hline
    \end{tabular}
    }
    \caption{The percent(\%) of score (0/1/2) of Knowledge Relevance and Context Coherence for Wizard Transformer, ITE+CKAD and ITE+DD.} \vspace{-10pt}
    \label{Score Distribution}
\end{table}

\subsection{Experiment Setup}
We use OpenNMT-py\footnote{\url{https://github.com/OpenNMT/OpenNMT-py}} \cite{opennmt} as the code framework\footnote{The code and models are available at \url{https://github.com/lizekang/ITDD}}. For all models, the hidden size is set to 512. For rnn-based models (Seq2Seq, HRED), 3-layer bidirectional LSTM \cite{hochreiter1997long} and 1-layer LSTM is applied for encoder and decoder respectively. For transformer-based models, the layers of both encoder and decoder are set to 3. The number of attention heads in multi-head attention is 8 and the filter size is 2048. The word embedding is shared by utterances, knowledge and generated responses. The dimension of word embedding is set to 512 empirically. We use Adam \cite{kingma2014adam} for optimization. When decoding, beam size is set to 5. We use the previous three utterances and its related documents as input.

\begin{table*}[!t]
\centering
\scalebox{0.95} {
\begin{tabular}{|rp{13cm}|}
  \hline
  \multicolumn{2}{|p{15cm}|}{\textbf{Document}} \\ \hline
  \multicolumn{2}{|p{15cm}|}{moviename \underline{\textbf{despicable me}} ...  \underline{\textbf{cast: steve carell as gru}} ... \underline{\textbf{rotten tomatoes: 81\%}} ... it is the debut film of \underline{\textbf{illumination entertainment}}. ... gru, a \underline{\textbf{supervillain}}, is disheartened when an unknown supervillain steals the great pyramid of giza. gru, with the assistance of his colleague dr. Nefario and his \underline{\textbf{minions}}, resolves to one-up this mystery rival by shrinking and stealing the moon. as it would be costly to steal the moon, \underline{\textbf{gru seeks a loan from the bank of evil}}. ... upon learning that an up-and-coming \underline{\textbf{villain}} named vector was responsible for the pyramid theft ...} \\ 
  \hline
  \textbf{Utterance 1} & what do you think about \underline{\textbf{despicable me}}? \\ 
  \hline
  \textbf{ITE+DD} & i think it is an awesome movie, \underline{\textbf{what about you?}} \\ 
  \textbf{Wizard} & i think it's a great movie! \\ 
  \hline
  \textbf{Utterance 2} & i rather liked it. it wasn't what i was expecting but it was still good. \\ 
  \hline
  \textbf{ITE+DD} & yeah, \underline{\textbf{it makes me laugh}}, even when they make a movie about it. \\ 
  \textbf{Wizard} & it's a great movie for kids and adults. \\ 
  \hline
  \textbf{Utterance 3} & i like it too. \underline{\textbf{funny and heart warming at the same time}}. \\ 
  \hline
  \textbf{ITE+DD} & i think it's interesting that \underline{\textbf{gru decided to take a loan from the bank of evil}}. \\ 
  \textbf{Wizard} & i think they should have made a sequel. \\ 
  \hline
  \textbf{Utterance 4} & \underline{\textbf{gru was a really interesting character}}, too. \underline{\textbf{i liked steve carell and the voice he}} \underline{\textbf{used for gru}}. \\ 
  \hline
  \textbf{ITE+DD} & yeah, now that you mention it, what did you think of the \underline{\textbf{minions}}? \\ 
  \textbf{Wizard} & i think the entire cast was great. \\ 
  \hline
  \textbf{Utterance 5} & yes, he did an excellent job portraying the character. \underline{\textbf{81\% in ratings}} i think is great. \\ 
  \hline
  \textbf{ITE+DD} & yes, \underline{\textbf{81\%}} is a great score! \\ 
  \textbf{Wizard} & i also liked the \underline{\textbf{villain}} of the movie. \\ 
  \hline
  
\end{tabular}
}
\caption{Responses generated by Incremental Transformer with Deliberation Decoder(\textbf{ITE+DD}) and the Wizard Transformer(\textbf{Wizard}). These utterances are from continuous dialogues. \textbf{ITE+DD} and \textbf{Wizard} generate responses using context utterances and a given document. Note that \textbf{Utterance $i+1$} is the gold response of \textbf{Utterance $i$}.} \vspace{-10pt}
\label{Case Study}
\end{table*}

\subsection{Evaluation Metrics}

\textbf{Automatic Evaluation}:
We adopt perplexity (PPL) and BLEU \cite{Papineni02bleu:a} to automatically evaluate the response generation performance. 
Models are evaluated using perplexity of the gold response as described in \cite{dinan2018wizard}. Lower perplexity indicates better performance. 
BLEU measures n-gram overlap between a generated response and a gold response. However, since there is only one reference for each response and there may exist multiple feasible responses, BLEU scores are extremely low. We compute BLEU score by the \emph{multi-bleu.perl}\footnote{\url{https://github.com/google/seq2seq/blob/master/bin/tools/multi-bleu.perl}} \\
\textbf{Manual Evaluation}:
Manual evaluations are essential for dialogue generation. We randomly sampled 30 conversations containing 606 utterances from the test set and obtained 5454 utterances from the nine models. We have annotators score these utterances given its previous utterances and related documents.
We defined three metrics - \textbf{fluency}, \textbf{knowledge relevance} \cite{liu2018knowledge} and \textbf{context coherence} for manual evaluation. All these metrics are scored 0/1/2. 

\textbf{fluency}: Whether the response is natural and fluent. Score 0 represents not fluent and incomprehensible; 1 represents partially fluent but still comprehensible; 2 represents totally fluent.

\textbf{knowledge relevance}: Whether the response uses relevant and correct knowledge. Score 0 represents no relevant knowledge; 1 represents containing relevant knowledge but not correct; 2 represents containing relevant knowledge and correct. 

\textbf{context coherence}: Whether the response is coherent with the context and guides the following utterances. Score 0 represents not coherent or leading the dialogue to an end; 1 represents coherent with the utterance history but not guiding the following utterances; 2 represents coherent with utterance history and guiding the next utterance.

\subsection{Experimental Results}

Table \ref{Automatic Evaluation and Munual Evaluation} shows the automatic and manual evaluation results for both the baseline and our models. 

In manual evaluation, among baselines, Wizard Transformer and RNN without knowledge have the highest fluency of 1.62 and Wizard obtains the highest knowledge relevance of 0.47 while Transformer without knowledge gets the highest context coherence of 0.67. For all models, ITE+CKAD obtains the highest fluency of 1.68 and ITE+DD has the highest Knowledge Relevance of 0.56 and highest Context Coherence of 0.90.

In automatic evaluation, our proposed model has lower perplexity and higher BLEU scores than baselines. For BLEU, HRED with knowledge obtains the highest BLEU score of 0.77 among the baselines. And ITE+DD gets 0.95 BLEU score, which is the highest among all the models. For perplexity, Wizard Transformer obtains the lowest perplexity of 70.30 among baseline models and ITE+DD has remarkably lower perplexity of 15.11 than all the other models. A detailed analysis is in Section \ref{Analysis}.

\subsection{Analysis and Discussion} \label{Analysis}
To our surprise, ITE+DD reaches an extremely low ground truth perplexity. We find that the ground truth perplexity after the first-pass decoding is only similar to the ITE+CKAD. It shows that the second-pass decoder utilizes the document knowledge well, and dramatically reduced the ground truth perplexity.

As shown in Table \ref{Score Distribution}, ITE+DD has a higher percent of score 2 both on Knowledge Relevance and Context Coherence than ITE+CKAD. This result also demonstrates that Deliberation Decoder can improve the knowledge correctness and guide the following conversations better.

Although the perplexity of ITE+CKAD is only slightly better than KAT, the BLEU score, Fluency, Knowledge Relevance and Context Coherence of ITE+CKAD all significantly outperform those of KAT model, which indicates that Incremental Transformer can deal with multi-turn document grounded conversations better. 

Wizard Transformer has a great performance on Knowledge Relevance only second to our proposed Incremental Transformer. However, its score on Context Coherence is lower than some other baselines. As shown in Table \ref{Score Distribution}, Wizard Transformer has Knowledge Relevance score 1 results twice more than score 2 results, which indicates that the model tends to generate responses with related knowledge but not correct. And the poor performance on Context Coherence also shows Wizard Transformer does not respond to the previous utterance well. This shows the limitation of representing context and document knowledge by simple concatenation.

\begin{table}[t]
\small
\centering
\scalebox{0.95} {
\begin{tabular}{|c|p{3cm}|p{3cm}|}
    \hline
    \textbf{ID} & \textbf{Utterance} &  \textbf{Two-pass Responses} \\
    \hline
    1 & \multirow{2}{3cm}{I think rachel mcadams had an even better role as regina george however! would you agree?} & i'm not a fan of \underline{\textbf{kristen bell}}, but i think \underline{\textbf{she}} did a great job. \\ \cline{3-3}
    & & i'm not a huge fan of \underline{\textbf{rachel mcadams}}, but \underline{\textbf{he}} did a great job. \\
    
    \hline
    2 & \multirow{2}{3cm}{yeah, I guess that's always worth it, and a truce was made as well.} & yeah, \underline{\textbf{not only}} does she reconcile with the \underline{\textbf{plastics}}. \\ \cline{3-3}
    & & yeah, she reconciles with \underline{\textbf{janis}} , \underline{\textbf{damien}} and \underline{\textbf{aaron}}. \\
    \hline
    3 & \multirow{2}{3cm}{i liked the scene where buzz thinks he's a big shot hero but then the camera reveals him to be a tiny toy.} & i think that's one of the best scenes in the movie. \\ \cline{3-3}
    & & oh, i think that is what makes the movie unique as well. \underline{\textbf{have}} \underline{\textbf{you seen any of the}} \underline{\textbf{other pixar movies}}? \\
    \hline
    
\end{tabular}
}
\caption{Examples of the two pass decoding. Underlined texts are the differences between two results. For each case, the first-pass response is on the top.} \vspace{-10pt}
\label{examples}
\end{table}

\subsection{Case Study}
In this section, we list some examples to show the effectiveness of our proposed model.

Table \ref{Case Study} lists some responses generated by our proposed Incremental Transformer with Deliberation Decoder (ITE+DD) and Wizard Transformer (which achieves overall best performance among baseline models). Our proposed model can generate better responses than Wizard Transformer on knowledge relevance and context coherence.

To demonstrate the effectiveness of the two-pass decoder, we compare the results from the first-pass decoding and the second-pass decoding. Table \ref{examples} shows the improvement after the second-pass decoding.  For Case 1, the second-pass decoding result revises the knowledge error in the first-pass decoding result. For Case 2, the second-pass decoder uses more detailed knowledge than the first-pass one. For Case 3, the second-pass decoder cannot only respond to the previous utterance but also guide the following conversations by asking some knowledge related questions.

\section{Related Work}
The closest work to ours lies in the area of open-domain dialogue system incorporating unstructured knowledge. 
~\citeauthor{ghazvininejad2018knowledge}~\shortcite{ghazvininejad2018knowledge} uses an extended Encoder-Decoder where the decoder is provided with an encoding of both the context and the external knowledge. 
~\citeauthor{parthasarathi2018extending}~\shortcite{parthasarathi2018extending} uses an architecture containing a Bag-of-Words Memory Network fact encoder and an RNN decoder.
~\citeauthor{dinan2018wizard}~\shortcite{dinan2018wizard} combines Memory Network architectures to retrieve, read and condition on knowledge, and Transformer architectures to provide text representation and generate outputs.
Different from these works, we greatly enhance the Transformer architectures to handle the document knowledge in multi-turn dialogue from two aspects: 1) using attention mechanism to combine document knowledge and context utterances; and 2) exploiting incremental encoding scheme to encode multi-turn knowledge aware conversations.

Our work is also inspired by several works in other areas. 
~\citeauthor{zhang2018improving}~\shortcite{zhang2018improving} introduces document context into Transformer on document-level Neural Machine Translation (NMT) task. 
~\citeauthor{guan2018story}~\shortcite{guan2018story} devises the incremental encoding scheme based on rnn for story ending generation task. 
In our work, we design an Incremental Transformer to achieve a knowledge-aware context representation using an incremental encoding scheme. 
~\citeauthor{xia2017deliberation}~\shortcite{xia2017deliberation} first proposes Deliberation Network based on rnn on NMT task. 
Our Deliberation Decoder is different in two aspects:
1) We clearly devise the two decoders targeting context and knowledge respectively; 2) Our second pass decoder directly fine tunes the first pass result, while theirs uses both the hidden states and results from the first pass.

\section{Conclusion and Future Work}
In this paper, we propose an Incremental Transformer with Deliberation Decoder for the task of Document Grounded Conversations. Through an incremental encoding scheme, the model achieves a knowledge-aware and context-aware conversation representation. By imitating the real-world human cognitive process, we propose a Deliberation Decoder to optimize knowledge relevance and context coherence.
Empirical results show that the proposed model can generate responses with much more relevance, correctness, and coherence compared with the state-of-the-art baselines. In the future, we plan to apply reinforcement learning to further improve the performance.

\section{Acknowledgments}
This work is supported by 2018 Tencent Rhino-Bird Elite Training Program, National Natural Science Foundation of China (NO. 61662077, NO.61876174) and National Key R\&D Program of China (NO.YS2017YFGH001428). We sincerely thank the anonymous reviewers for their thorough reviewing and valuable suggestions.

\bibliography{acl2019}
\bibliographystyle{acl_natbib}

\end{document}